%% file: mining_gold.tex
\documentclass{article}

\PassOptionsToPackage{numbers, sort&compress}{natbib}

\usepackage[preprint]{neurips_2019}

\usepackage[utf8]{inputenc}
\usepackage[T1]{fontenc}
\usepackage[british]{babel}
\usepackage[utf8]{inputenc}
\usepackage{color,xcolor}
\usepackage{wrapfig}

\definecolor{dark-red}{rgb}{0.8, 0.0, 0.1803921568627451}
\definecolor{dark-blue}{rgb}{0.0, 0.0, 0.803921568627451}
\definecolor{dark-green}{rgb}{0.0, 0.39215686274509803, 0.0}
\definecolor{dark-orange}{rgb}{0.8, 0.4, 0.0}

\usepackage{amsmath,amsfonts, dsfont,mathtools,}
\usepackage{graphicx}

\usepackage{xfrac}

\usepackage{subcaption}
\usepackage{tikz}
\usetikzlibrary{bayesnet}

\usepackage{multirow,booktabs,tabularx,float}
\PassOptionsToPackage{hyphens}{url}\usepackage[pdftex,hidelinks,linkcolor={dark-blue}, citecolor={dark-green}, urlcolor={dark-red},bookmarksopen,linktoc=all]{hyperref}
\usepackage{enumitem} 
\usepackage[stretch=15,shrink=15]{microtype} 
\usepackage{xspace} 
\usepackage{wrapfig}
\usepackage[font=small]{caption}

\input{definitions}

\begin{document}

\title{Mining gold from implicit models to  improve likelihood-free inference}

\author{Johann Brehmer,$^1$ Gilles Louppe,$^2$ Juan Pavez,$^3$ and Kyle Cranmer$^1$ \\
{}$^1$\,New York University, {}$^2$\,University of Li\`{e}ge, {}$^3$\,Federico Santa Mar\'ia Technical University\\
\texttt{johann.brehmer@nyu.edu}, \texttt{g.louppe@uliege.be},\\
\texttt{juan.pavezs@alumnos.usm.cl}, \texttt{kyle.cranmer@nyu.edu} \\
}

\maketitle

\begin{abstract}
Simulators often provide the best description of real-world phenomena. However, the density they implicitly define is often intractable, leading to challenging inverse problems for inference. Recently, a number of techniques have been introduced in which a surrogate for the intractable density is learned, including normalizing flows and density ratio estimators. We show that additional information that characterizes the latent process can often be extracted from simulators and used to augment the training data for these surrogate models. We introduce several new loss functions that leverage this augmented data, and demonstrate that these new techniques can improve sample efficiency and quality of inference.
\end{abstract}

\section{Introduction}

In many areas of science, complicated real-world phenomena are best described
through computer simulations. Typically, the simulators  implement a stochastic
generative process in the ``forward'' mode based on a well-motivated mechanistic
model with parameters $\theta$. While the simulators can generate samples of
observations $x \sim p(x | \theta)$, they typically do not admit a tractable
likelihood (or density) $p(x | \theta)$. Probabilistic models defined only via
the samples they produce are often called implicit models. Implicit models lead
to intractable inverse problems, which is a barrier for statistical inference of
the parameters $\theta$ given observed data. These problems arise in fields as
diverse as particle physics, epidemiology, and population genetics, which has
motivated the development of \emph{likelihood-free inference} algorithms such as
Approximate Bayesian Computation (\abc)~\cite{rubin1984,
beaumont2002approximate, Alsing:2018eau, Charnock:2018ogm} and neural density
estimation (\nde) techniques~\cite{2014arXiv1406.2661G, Cranmer:2015bka,
Cranmer:2016lzt, Louppe:2016aov, 2016arXiv161003483M, gutmann2017likelihood,
2018arXiv181009899D, Hermans:2019ioj, 2017arXiv170208896T,  2014arXiv1410.8516D,
2015arXiv150505770J, 2016arXiv160508803D, 2017arXiv170507057P,
2018arXiv180400779H, 2018arXiv180507226P, DBLP:journals/corr/abs-1806-07366,
2018arXiv180703039K, 2018arXiv181001367G, 2015arXiv150203509G,
2016arXiv160502226U, 2016arXiv160903499V, 2016arXiv160605328V,
2016arXiv160106759V, 2012arXiv1212.1479F, NIPS2016_6084, 2016arXiv160206701P,
2016arXiv161110242D, 2017arXiv170707113L, 2017arXiv171101861L,
2018arXiv180509294L}. While many of these techniques can be exact in the limit
of infinite training samples, real-world simulators are computationally
expensive, and sample efficiency is immensely important.

We present a suite of new techniques that can dramatically improve the sample
efficiency for training neural network surrogates that estimate the likelihood
$p(x | \theta)$ or likelihood ratio $r(x | \theta_0, \theta_1) = p(x | \theta_0)
/ p(x | \theta_1)$. This provides the key quantity needed for both frequentist
and Bayesian inference procedures. Our approach involves extracting additional
information that characterizes the latent process from the simulator, as we
explain in Sec.~\ref{sec:mining_gold}. In Sec.~\ref{sec:learning_from_gold} we
introduce the loss functions that utilize  this augmented data. In
Sec.~\ref{sec:examples} we demonstrate through a range of experiments that these
new techniques provide a significant increase in sample efficiency compared to
techniques that do not leverage the augmented data, which ultimately increase
the quality of inference.

\section{Related work}
\label{sec:related}

Techniques for likelihood-free inference can be divided into two broad
categories. In the first category, the inference is performed by directly
comparing the observed data to the output of the simulator. This includes
Approximate Bayesian Computation (\abc)~\cite{rubin1984,
beaumont2002approximate, Alsing:2018eau, Charnock:2018ogm} and probabilistic
programming systems~\cite{wood-aistats-2014,le2017inference}. Here we focus on a
second category, in which the simulator is used to generate training data for a
tractable surrogate model that is used during inference. There are  rich
connections between simulator-based inference and learning in implicit
generative models such as GANs, with a considerable amount of cross-pollination
between these areas~\citep{2016arXiv161003483M}.

\paragraph{The likelihood ratio trick (\lrt).}
A surrogate model for the likelihood ratio $\hat{r}(x|\theta_0, \theta_1)$ can
be defined by training a probabilistic classifier to discriminate between two
equal-sized samples $\{x_i\} \sim p(x | \theta_0)$ and $\{x_i\} \sim p(x |
\theta_1)$. The binary cross-entropy loss
\begin{equation}
  L_\textrm{XE} = -\mathbb{E}[\mathds{1}(\theta = \theta_1) \log \hat{s}(x | \theta_0, \theta_1)
  + \mathds{1}(\theta = \theta_0) \log (1-\hat{s}(x | \theta_0, \theta_1)) ]
  \label{eq:loss_crossentropy}
\end{equation}
is minimized by the optimal decision function $s(x| \theta_0, \theta_1) = p(x |
\theta_1) / (p(x | \theta_0) + p(x | \theta_1))$. Inverting this relation, the
likelihood ratio can be estimated from the classifier decision function
$\hat{s}(x)$ as $\hat{r}(x | \theta_0, \theta_1) = (1 - \hat{s}(x| \theta_0,
\theta_1))/\hat{s}(x| \theta_0, \theta_1)$. This ``likelihood ratio trick'' is
widely appreciated~\cite{2014arXiv1406.2661G, Cranmer:2015bka, Cranmer:2016lzt,
Louppe:2016aov, 2016arXiv161003483M, gutmann2017likelihood, 2018arXiv181009899D,
Hermans:2019ioj, 2017arXiv170208896T}. In practice, not all probabilistic
classifiers trained to separate samples from $\theta_0$ and $\theta_1$ learn the
optimal decision function. As long as the classifier decision function is a
monotonic function of the likelihood ratio, this relation can be restored
through a calibration procedure, substantially increasing the applicability of
the likelihood ratio trick~\cite{Cranmer:2015bka, Cranmer:2016lzt,
Louppe:2016aov}. We use the term \carl ({\textbf{C}alibrated
\textbf{a}pproximate \textbf{r}atios of \textbf{l}ikelihoods}) to describe
likelihood ratio estimators based on calibrated classifiers.

\paragraph{Neural density estimation (\nde).}
More recently, several methods for conditional density estimation have been
proposed, often based on neural networks~\cite{2014arXiv1410.8516D,
2015arXiv150505770J, 2016arXiv160508803D, 2017arXiv170507057P,
2018arXiv180400779H, 2018arXiv180507226P, DBLP:journals/corr/abs-1806-07366,
2018arXiv180703039K, 2018arXiv181001367G, 2015arXiv150203509G,
2016arXiv160502226U, 2016arXiv160903499V, 2016arXiv160605328V,
2016arXiv160106759V, 2012arXiv1212.1479F, NIPS2016_6084, 2016arXiv160206701P,
2016arXiv161110242D, 2017arXiv170208896T, 2017arXiv170707113L,
2017arXiv171101861L, 2018arXiv180509294L}. They can be used to train a surrogate
for the likelihood $p(x|\theta)$~\cite{2017arXiv170507057P, 2018arXiv180507226P,
2018arXiv180509294L} or, in a Bayesian setting, the posterior
$p(\theta|x)$~\cite{NIPS2016_6084, 2017arXiv171101861L, 2019arXiv190507488G}.
One particularly interesting class of models are normalizing
flows~\cite{2014arXiv1410.8516D, 2015arXiv150505770J, 2016arXiv160508803D,
2017arXiv170507057P, 2018arXiv180400779H, 2018arXiv180507226P,
DBLP:journals/corr/abs-1806-07366, 2018arXiv180703039K, 2018arXiv181001367G},
which model the density as a sequence of invertible transformations applied to a
simple base density. The target density is then given by the Jacobian
determinant of the transformation. Closely related, autoregressive
models~\cite{2015arXiv150203509G, 2016arXiv160502226U, 2016arXiv160903499V,
2016arXiv160605328V, 2016arXiv160106759V} factorize a target density as a
sequence of simpler conditional densities.

\paragraph{Novel contributions.}
The most important novel contribution that differentiates our work from the
existing methods is the observation that additional information can be extracted
from the simulator, and the development of loss functions that allow us to use
this ``augmented'' data to more efficiently learn surrogates for the likelihood
function. In addition, we show how the augmented data can be used to define
locally optimal summary statistics, which can then be used for inference with
density estimation techniques or \abc. We playfully introduce the analogy of
mining gold as this augmented data requires work to extract and is very
valuable. In experiments we demonstrate that these approaches can dramatically
improve sample efficiency and quality of likelihood-free inference.

Concurrently, the application of these methods to a specific class of problems
in particle physics has been discussed in Refs.~\cite{Brehmer:2018kdj,
Brehmer:2018eca}. The present manuscript is meant to serve as the primary
reference for these new techniques and is addressed to the broader physical
science and machine learning communities. Most importantly, it generalizes the
specific particle physics case to almost any scientific simulator, requiring
significantly weaker assumption than those made in the physics context. We also
introduce an entirely new algorithm called \scandal, for which we provide the
first experimental results.

\section{Extracting more information from the simulator}
\label{sec:mining_gold}

We consider a scientific simulator that implements a stochastic generative
process that proceeds through a series of latent states $z_i \in \mathcal{Z}_i$
and finally to an output $x \in \mathbb{R}^{d_x}$. The latent space structure
$\mathcal{Z}$ can involve discrete and continuous components and is derived from
the control flow of the (differentiable or non-differentiable) simulation code.
Based on the mechanistic model implemented by the simulator, each latent state
is sampled from a conditional probability density $z_i \sim p_i(z_i | \theta,
z_{<i})$ and the final output is sampled from $x \sim p_x(x | \theta, z)$. The
likelihood is then given by
\begin{equation}
  p(x | \theta) = \intz p(x, z | \theta)
  =  \intz  p_x(x | \theta, z)  \;
  \prod_{i}  p_i(z_i | \theta, z_{<i})
  \,.
\end{equation}
Often the likelihood is intractable exactly because the latent space $z$ is
enormous and it is unfeasible to explicitly calculate this integral.  In
real-world scientific simulators, the trajectory for a single observation can
involve many millions of latent variables.

In this paper we consider the problem of estimating the likelihood $p(x|\theta)$
or the likelihood ratio $r(x|\theta_0, \theta_1)$, which for the practical
purpose of inferring parameter values $\theta$ can be used almost
interchangably, based on the data available from $N$ runs of the simulator.

Typically, the setting of likelihood-free inference assumes that the only
available output from the simulator are samples of observations  $x \sim p(x |
\theta)$. But in real-life simulators, more information can usually be
extracted. We typically have access to the latent variables $z$, and the
distributions of each latent variable $p_i(z_i | \theta, z_{<i})$ and
$p_x(x|\theta, z)$ are tractable.

The key observation that is the starting point of our new inference methods is
the following: While $p(x|\theta)$ is intractable, for each simulated sample we
can calculate the \emph{joint score}
\begin{equation}
  t(x , z | \theta_0) \equiv \nabla_{\theta} \log p(x, z | \theta) \biggr |_{\theta_0}
  = \sum_i \nabla_\theta \log p_i(z_i | \theta, z_{<i}) \biggr |_{\theta_0} + \nabla_\theta \log p_x(x | \theta, z)  \biggr |_{\theta_0}
  \label{eq:joint_score}
\end{equation}
by accumulating the factors $\nabla_{\theta} \log p(z_i | \theta, z_{:i})$ as
the simulation runs forward through its control flow conditioned on the random
trajectory $z$.  It can be insightful to think of the mechanistic model in the simulator as defining a policy $\pi_\theta$ and $t(x , z | \theta_0)$ as analogous to  the policy gradient
used in \toolfont{Reinforce}~\cite{williams1992simple}. However, instead of
trying to optimize $\theta$ via a stochastic gradient estimate of some reward
function, we will simply augment the data generated by the simulator with the
joint score.

Similarly, we can extract the \emph{joint likelihood ratio}
\begin{equation}
  r(x , z | \theta_0, \theta_1) \equiv \frac {p(x, z | \theta_0)} {p(x, z | \theta_1)}
  = \frac {p_x(x | \theta_0, z)} {p_x(x| \theta_1, z)} \;
  \prod_i \frac {p_i(z_i | \theta_0, z_{<i})} {p_i(z_i | \theta_1, z_{<i})} \,.
\end{equation}
The joint score and joint likelihood ratio quantify how much more or less likely
a particular simulated trajectory through the simulator would be if one changed
$\theta$.

 \begin{wrapfigure}{r}{0.4\textwidth}
   \centering
   \vspace*{-12pt}
   \includegraphics[width=0.4\textwidth,clip,trim=0.2cm 0.5cm 0 0.3cm]{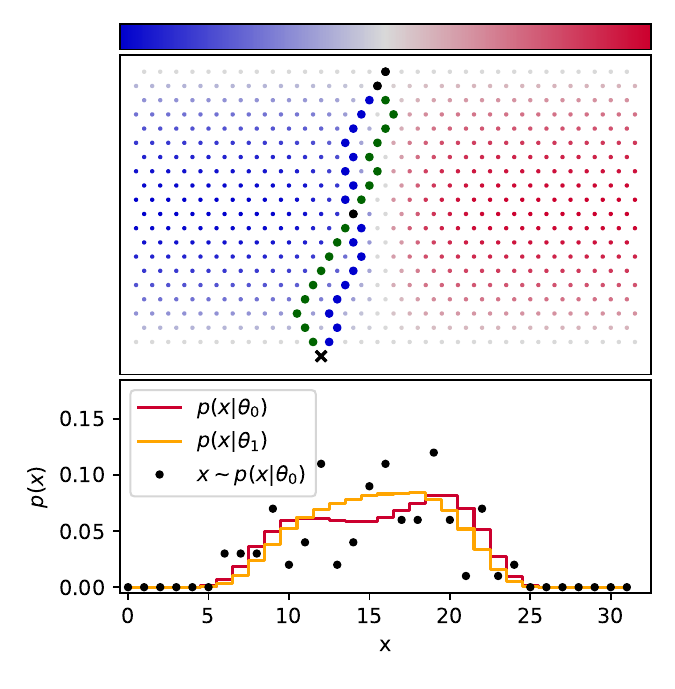}%
   \vspace*{-4pt}
   \caption{A toy simulation generalizing the Galton board where the transitions
   are biased left (blue) or right (red) depending on the nail position and the
   value of $\theta$. Two example latent trajectories $z$ are shown (blue and
   green), leading to the same observed value of $x$. Below, the distribution for
   $\theta_0= -0.8$ and $\theta_1 = -0.6$. An example empirical
   distribution from 100 runs for $\theta_0$ shows that the sample variance is much
   larger than the differences from $\theta_0$ vs $\theta_1$.}
   \label{fig:plinko}
   \vspace*{-12pt}
 \end{wrapfigure}

As a motivating example, consider the simulation for a generalization of the
Galton board, in which a set of balls is dropped through a lattice of nails
ending in one of several bins denoted by $x$. The Galton board is commonly used
to demonstrate the central limit theorem, and if the nails are uniformly placed
such that the probability of bouncing to the left is $p$, the sum over the
latent space is tractable analytically and the resulting distribution of $x$ is
a binomial distribution. However, if the nails are not uniformly placed, and the
probability of bouncing to the left is an arbitrary function of the nail
position and some parameter $\theta$, the resulting distribution requires an
explicit sum over the latent paths $z$ that might lead to a particular $x$. Such
a distribution would become intractable as the size of the lattice increases.
Figure~\ref{fig:plinko} shows  an example of two latent trajectories that lead
to the same $x$. In this toy example, the probability $p(z_h, z_v, \theta)$ of
going left is given by $(1 - f(z_v) )/2 + f(z_v)  {\sigma}( 5 \theta
(z_h-\sfrac{1}{2}))$, where $f(z_v) = \sin(\pi z_v)$, $\sigma$ is the sigmoid
function, and $z_h$ and $z_v$ are the horizontal and vertical nail positions
normalized to $[0,1]$. This leads to a non-trivial $p(x|\theta)$, which can even
be bimodal. Code for simulation and inference in this problem is available at
Ref.~\cite{repository-plinko}.

Figure~\ref{fig:plinko} shows that a large number of samples from the simulator
are needed to reveal the differences in the distribution of $x$ for small
changes in $\theta$ -- the number of samples needed grows like $(p/\Delta p)^2$.
Moreover, this toy simulation is representative of many real-world simulators in
that it is composed of  non-differentiable control-flow elements. This poses a
difficulty, making methods based on $\nabla_z x$~\cite{graham2017asymptotically}
and $\nabla_\theta x$ inapplicable, which previously motivated techniques such
as Adversarial Variational Optimization~\cite{2017arXiv170707113L}. But the
joint score in Eq.~\eqref{eq:joint_score} can easily be computed by accumulating
the factors $\nabla_{\theta} \log p(z_h, z_v | \theta_0)$, and we can calculate
the joint likelihood ratio by accumulating factors $p(z_h, z_v | \theta_0) /
p(z_h, z_v | \theta_1)$. In analogy to the Galton board toy example, even
complicated real-world simulators often allow us to accumulate these factors as
they run, and to calculate the joint score and joint likelihood ratio
conditional on a particular stochastic execution trace $z$. We will demonstrate
this with two more examples in Sec.~\ref{sec:examples}.

For simulators written in an automatic differentiation framework, the
calculation of the joint score and joint likelihood ratio can be entirely
automatic and does not require any changes to the simulator code and output. As
a proof of principle, at Ref.~\cite{repository-automization} we provide a
framework that automates these calculations for any simulator in which all
stochastic steps are implemented with the \toolfont{Pyro}
library~\cite{bingham2018pyro}.

\section{Learning from augmented data}
\label{sec:learning_from_gold}

\subsection{Key idea}

How can the ``augmented data'' consisting of simulated observations ${x_i}$, the
joint likelihood ratio $r(x_i, z_i | \theta_0, \theta_1)$, and the joint score
$t(x_i, z_i | \theta_0)$ be used to estimate the likelihood $p(x|\theta)$ or
likelihood ratio $r(x | \theta_0, \theta_1)$? The relation between $r(x,z |
\theta_0, \theta_1)$ and $r(x | \theta_0, \theta_1)$ is not trivial\,---\,the
integral of the ratio is not the ratio of the integrals! Similarly, how can the
joint score be used to estimate the intractable score function
\begin{equation}
  t(x | \theta_0) \equiv \nabla_{\theta} \log p(x | \theta) \biggr |_{\theta_0} \,?
  \label{eq:score}
\end{equation}
The integral of the log is not the log of the integral!

Consider the squared error of a function $\hat{g}(x)$ that only depends on the
observable $x$, but is trying to approximate a function $g(x,z)$ that also
depends on the latent variable $z$,
\begin{equation}
  L_\textrm{MSE} = \mathbb{E}_{p(x,z|\theta)} \left [ \;  \left ( g(x,z) - \hat{g}(x) \right )^2 \, \right ] .
\end{equation}
The minimum-mean-squared-error prediction of $\hat{g}(x)$ is given by the
conditional expectation $g^*(x) = \argmin_{\hat{g}} L_\textrm{MSE} =
\mathbb{E}_{p(z|x,\theta)} [g(x,z)]$.

Identifying $g(x, z)$ with the joint likelihood ratio $r(x,  z | \theta_0,
\theta_1)$ and $\theta = \theta_1$, we define
\begin{equation}
  L_r = \mathbb{E}_{p(x,z|\theta_1)} \left [ \;  \left ( r(x,z|\theta_0, \theta_1) - \hat{r}(x) \right )^2 \, \right ] \,,
  \label{eq:loss_functional_ratio}
\end{equation}
and find that this functional is minimized by $r^*(x) = \argmin_{\hat{r}} L_r =
\mathbb{E}_{p(z|x, \theta_1)} \left [ \, r(x,z|\theta_0, \theta_1) \, \right ] =
r(x | \theta_0, \theta_1)$. Similarly, by identifying $g(x, z)$ with the joint
score $t(x,  z | \theta_0)$ and setting $\theta = \theta_0$, we define
\begin{equation}
  L_t = \mathbb{E}_{p(x,z|\theta_0)} \left [ \;  \left ( t(x,z|\theta_0) - \hat{t}(x|\theta_0) \right )^2 \, \right ] \,,
  \label{eq:Lt}
\end{equation}
which is minimized by $t^*(x) =  \mathbb{E}_{p(z|x, \theta_0)} \left [ \,
t(x,z|\theta_0) \, \right ] = t(x | \theta_0)$.

These loss functionals are immensely useful because they allow us to transform
$t(x,z | \theta_0)$ into $t(x | \theta_0)$ and $r(x,z| \theta_0, \theta_1)$ into
$r(x| \theta_0, \theta_1)$: we are able to regress on these two intractable
quantities! This is what makes the joint score and joint likelihood ratio the
gold worth mining.

\subsection{Learning the likelihood ratio}
\label{sec:learning_likelihood_ratios}

Based on this observation we introduce a family of new likelihood-free inference
techniques. They fall into two categories. We first discuss a class of
algorithms that uses the augmented data to learn a surrogate model for any
likelihood $p(x|\theta)$ or likelihood ratio $r(x | \theta_0, \theta_1)$. In
Section~\ref{sec:local_model} we will define a second class of methods that is
based on a local expansion of the model around some reference parameter point.

The simulators we consider in this work do not only implicitly define a single
density $p(x)$, but a family of densities $p(x|\theta)$. The parameters $\theta$
may potentially belong to a high-dimensional parameter space. For inference
models based on surrogate models, there are two broad strategies to model this
dependence. The first is to estimate $p(x|\theta)$ or the likelihood ratio
$r(x|\theta_0, \theta_1)$ for specific values of $\theta$ or pairs $(\theta_0,
\theta_1)$. This may be done via a pre-defined set of $\theta$ values or
on-demand using an active-learning iteration. We follow a second approach, in
which we train \emph{parameterized estimators} for the full model $\hat{p}(x |
\theta)$ or $\hat{r}(x | \theta_0,  \theta_1)$ as a function of both the
observables $x$ and the parameters $\theta$~\cite{Cranmer:2015bka,
Baldi:2016fzo}. The training data then consists of a number of samples, each
generated with different values of $\theta_0$ and $\theta_1$, and the parameter
values are used as additional inputs to the surrogate model. When modeling the
likelihood ratio, the reference hypothesis $\theta_1$ in the denominator of the
likelihood ratio can be kept at a fixed reference value (or a marginal model
with some prior $\pi(\theta_1)$), and only the $\theta_0$ dependence is modeled
by the network. This approach encourages the estimator to learn the typically
smooth dependence of the likelihood ratio on the parameters of interest from the
training data and borrow power from neighboring points.

\paragraph{\rolr (Regression On Likelihood Ratio):}
First, a number of parameter points $(\theta_0, \theta_1)$ is drawn from
$\theta_i \sim \pi_i(\theta_i)$.  For each pair $(\theta_0, \theta_1)$, we run
the simulator both for $\theta_0$ and for $\theta_1$, labelling the samples with
$y = 0$ and $y=1$, respectively. In addition to samples $x \sim p(x | \theta_y)$
we also extract the joint likelihood ratio $r(x,z|\theta_0, \theta_1)$.

An expressive regressor $\hat{r}(x | \theta_0, \theta_1)$ (\eg a neural network)
is trained by minimizing the squared error loss
\begin{equation}
  L_{\rolr}[\hat{r}] = \frac 1 N \sum_i \Biggl( y_i \, |r(x_i,z_i) - \hat{r}(x_i)|^2
    + (1-y_i)  \, \left |\frac 1 {r(x_i,z_i)} - \frac 1 {\hat{r}(x_i)} \right|^2 \Biggr) .
\end{equation}
Here and in the following the $\theta$ dependence is implicit to reduce the
notational clutter.

Both terms in this loss function are estimators of
Eq.~\eqref{eq:loss_functional_ratio} (in the second term we switch $\theta_0
\leftrightarrow \theta_1$ to reduce the variance by mapping out other regions of
$x$ space). As we showed in the previous section, this loss function is, at
least in the limit of infinite data, minimized by the true likelihood ratio $r(x
| \theta_0, \theta_1)$. A regressor trained in this way thus provides an
estimator for the likelihood ratio and can be used for frequentist or Bayesian
inference methods.

\paragraph{\rascal (Ratio And SCore Approximate Likelihood ratio):}
If such a likelihood ratio regressor is differentiable (as is the case for
neural networks) with respect to $\theta$, we can calculate the predicted score
$\hat{t}(x | \theta_0) = \nabla_{\theta_0} \log \hat{r} (x | \theta_0, \theta_1
)$.  For a perfect likelihood ratio estimator, $\hat{t}(x | \theta_0)$ minimizes
the squared error with respect to the joint score, see Eq.~\eqref{eq:Lt}.
Turning this argument around, we can improve the training of a likelihood ratio
estimator by minimizing the combined ratio and score loss with a hyper-parameter
$\alpha$
\begin{equation}
  L_{\rascal}[\hat{r}] =   L_{\rolr}[\hat{r}]
  + \alpha \frac 1  N \sum_i (1-y_i)  \left |t(x_i,z_i) -  \nabla_{\theta_0} \log \hat{r} (x_i)  \right|^2
  \; .
\end{equation}

\paragraph{\cascal (\carl And SCore Approximate Likelihood ratio):}
The same trick can be used to improve the likelihood ratio trick and the \carl
inference method~\cite{Cranmer:2015bka, Louppe:2016aov}. Following the
discussion around Eq.~\eqref{eq:loss_crossentropy}, a calibrated classifier
trained to discriminate samples $\{x_i\} \sim p(x | \theta_0)$ and $\{x_i\} \sim
p(x | \theta_1)$ provides a likelihood ratio estimator. For a differentiable
parameterized classifier, we can calculate the surrogate score $\hat{t}(x |
\theta_0) = \nabla_{\theta_0} \log [(1 - \hat{s}(x| \theta_0, \theta_1)) /
\hat{s}(x| \theta_0, \theta_1)]$. This allows us to train an improved classifier
(and thus a likelihood ratio estimator) by minimizing the combined loss
\begin{equation}
  L_{\cascal}[\hat{s}] =   L_{\mathrm{XE}}[\hat{s}]
  + \alpha \frac 1  N \sum_i (1-y_i)  \left |t(x_i,z_i) - \nabla_{\theta_0} \log \left[\frac {1 - \hat{s}(x)} {\hat{s}(x) }\right]  \right|^2 \,.
\end{equation}

\paragraph{\scandal (SCore-Augmented Neural Density Approximates Likelihood):}
Finally, we can use the same strategy to improve conditional neural density
estimators such as density  networks or normalizing flows. If a parameterized
neural density estimator $\hat{p}(x | \theta)$ is differentiable with respect to
$\theta$, we can calculate the surrogate score $\hat{t}(x) = \nabla_{\theta}
\log \hat{p}(x | \theta)$ and train an improved density estimator by minimizing
\begin{equation}
  L_{\scandal} [ \hat{p}] = L_{\mathrm{MLE}}
+ \alpha \frac 1  N \sum_i \left | t(x_i,z_i) - \nabla_{\theta} \log \hat{p}(x) \right|^2 \,.
\end{equation}
Unlike the methods discussed before, this provides an estimator of the
likelihood itself rather than its ratio. Depending on the architecture, the
surrogate also provides a generative model.

\subsection{Locally sufficient statistics for implicit models}
\label{sec:local_model}

A second class of new likelihood-free inference methods is based on an expansion
of the implicit model around a reference parameter point $\theta_{\text{ref}}$.
Up to linear order in $\theta - \theta_{\text{ref}}$, we find
\begin{equation}
  \localmodel(x|\theta) = \frac 1 {Z(\theta)} \, p(t(x|\theta_{\text{ref}})\,|\,\theta_{\text{ref}}) \, \exp[ t(x|\theta_{\text{ref}}) \cdot (\theta - \theta_{\text{ref}}) ]
  \label{eq:local_model}
\end{equation}
with some normalization factor $Z(\theta)$. This local approximation is in the
exponential family and the score vector $t(x | \theta_{\text{ref}})$, defined in
Eq.~\eqref{eq:score}, are its sufficient statistics.

For inference in a sufficiently small neighborhood around a reference point
$\theta_{\text{ref}}$, a precise estimator of the score $\hat{t}(x |
\theta_{\text{ref}})$ therefore defines a vector of ideal summary statistics
that contain all the information in an observation $x$ on the parameters
$\theta$~\citep[see also][]{Alsing:2017var, Alsing:2018eau, Alsing:2019dvb}. The
joint score together with a minimization of the loss in Eq.~\eqref{eq:Lt} allows
us to extract sufficient statistics from an intractable, non-differentiable
simulator, at least in the neighborhood of $\theta_\text{ref}$. Moreover, this
local model can be estimated by running the simulator at a single value
$\theta_\text{ref}$\,---\,it does not require scanning the $\theta$ space, and
thus avoids the curse of dimensionality. Based on this observation, we introduce
two further inference strategies:

\paragraph{\sally (Score Approximates Likelihood LocallY):}
By minimizing the squared error with respect to the joint score, see
Eq.~\eqref{eq:Lt}, we train a score estimator $\hat{t}(x |
\theta_{\text{ref}})$. In a next step, we estimate the density $\hat{p}(\hat{t}
(x | \theta_{\text{ref}}) | \theta)$ through standard multivariate density
estimation techniques. This calibration procedure implicitly includes the effect
of the normalizing constant $Z(\theta)$.

\paragraph{\sallino (Score Approximates Likelihood Locally IN One dimension):}
The \sally inference method requires density estimation in the estimated score
space, with typically $\dim \hat{t} \equiv \dim \theta \ll \dim x$. But in cases
with large number of parameters, it is beneficial to reduce the dimensionality
even further. In the local model of Eq.~\eqref{eq:local_model}, the likelihood
ratio $r(x | \theta_0, \theta_1)$ only depends on $h(x | \theta_0, \theta_1)
\equiv t(x | \theta_{\text{ref}}) \cdot (\theta_0 - \theta_1)$ up to an
$x$-independent constant related to $Z(\theta)$. Any neural score estimator lets
us also estimate this scalar function, which is a sufficient statistic for the
1-dimensional parameter space connecting $\theta_0$ and $\theta_1$. We can thus
estimate the likelihood ratio through univariate, rather than multivariate,
density estimation on $\hat{h}$.

The \sally and \sallino techniques are designed to work very well close to the
reference point. The local model approximation may deteriorate further away,
leading to a reduced sensitivity and weaker bounds. These approaches are simple
and robust, and in particular the \sallino algorithm scales exceptionally well
to high-dimensional parameter spaces.

For all these inference strategies, the augmented data is particularly powerful
for enhancing the power of simulation-based inference for small changes in the
parameter $\theta$. When restricted to samples $x \sim p(x|\theta)$, the
variance from the simulator is a challenge. The fluctuations in the empirical
density scale with the square root of the number of samples, thus large numbers
of samples are required before small changes in the implicit density can
faithfully be distinguished. In contrast, each sample of the joint ratio and
joint score provides an exact piece of information even for arbitrarily small
changes in $\theta$. On the other hand, the augmented data is less powerful for
deciding between model parameter points that are far apart. In this situation
the joint probability distributions $p(x,z|\theta)$ often do not overlap
significantly, and the joint likelihood ratio can have a large variance around
the intractable likelihood ratio $r(x|\theta_0, \theta_1)$. In addition, over
large distances in parameter space the local model is not valid and the score
does not characterize the likelihood ratios anymore, limiting the usefulness of
the joint score.

\section{Experiments}
\label{sec:examples}

\begin{figure}[t]
  \centering%
  \includegraphics[width=0.33\textwidth,clip,trim=0 0.5cm 0 0.4cm]{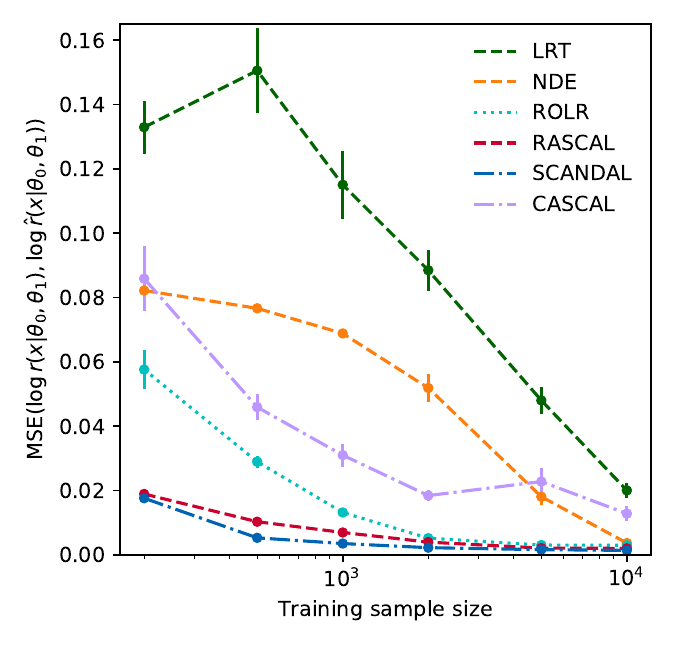}%
  \includegraphics[width=0.33\textwidth, clip,trim=0 0.5cm 0 0.3cm]{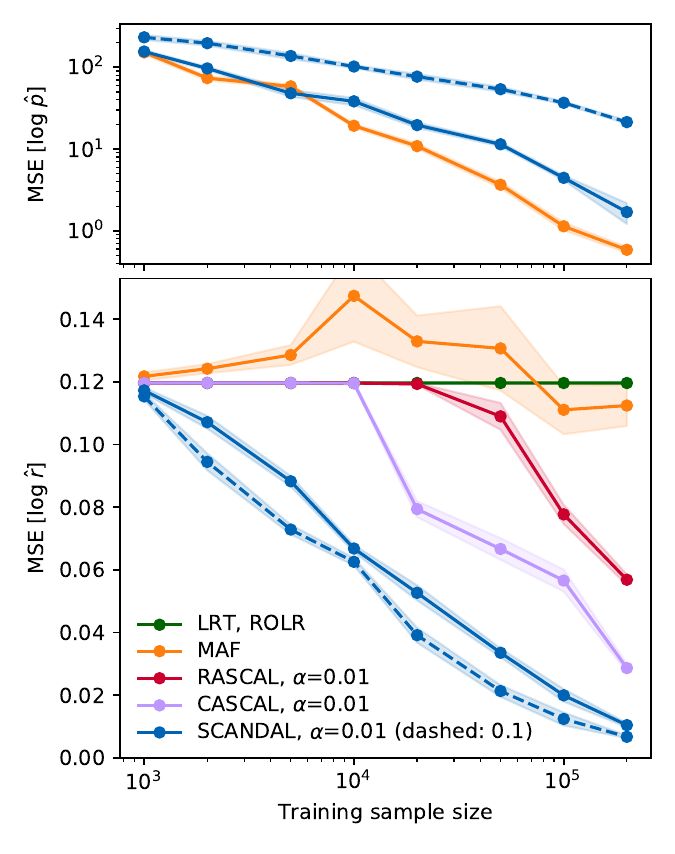}%
  \includegraphics[width=0.33\textwidth,clip,trim=0 0.5cm 0 0.4cm]{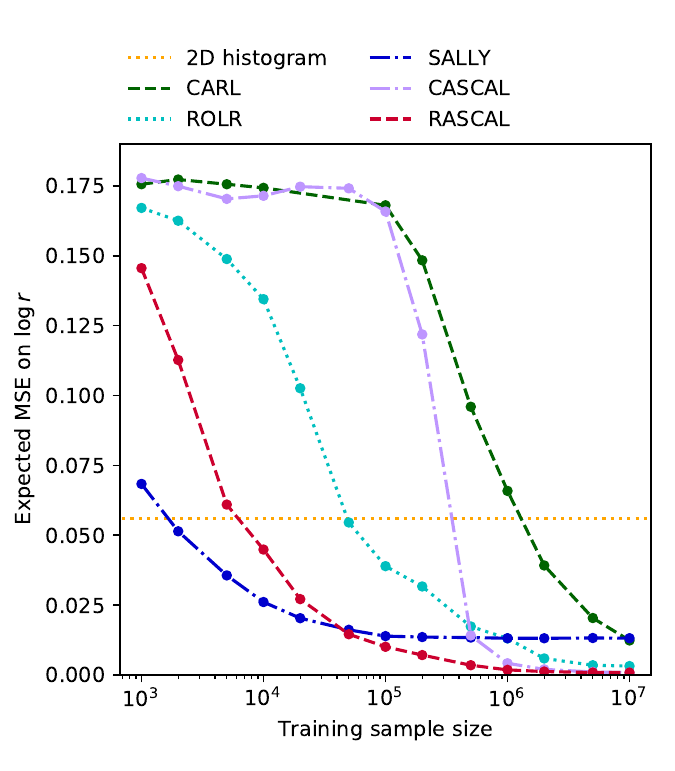}%
  \vspace*{-4pt}
  \caption{Left: Galton board example. MSE on $\log r$ vs.\ training sample size. We show the mean and its error based on 15 runs. Middle: Lotka-Volterra example. MSE on $\log p$ (top) and $\log r$ (bottom) vs.\ training sample size. We show the median and the standard deviation of 10 runs. Right: Particle physics example. MSE on $\log r$ vs.\ training sample size.}
  \label{fig:results}
  \vspace*{-20pt}
\end{figure}

\paragraph{Generalized Galton board.}
We return to the motivating example in Sec.~\ref{sec:mining_gold} and
Fig.~\ref{fig:plinko} and try to estimate likelihood ratios for the generalized
Galton board. We use the likelihood ratio trick and a neural density estimator
as baselines and compare them to the new \rolr, \rascal, \cascal, and \scandal
methods. As the simulator defines a distribution over a discrete $x$, for the
\nde and \scandal methods we use a neural network with a softmax output layer
over the bins to model $\hat{p}(x|\theta)$. All networks are explicitly
parameterized in terms of $\theta$, the parameter of the simulator that defines
the position of the nails (\ie they take $\theta$ as an input). We use a simple
network architecture with a single hidden layer, 10 hidden units, and $\tanh$
activations. The left panel of Fig.~\ref{fig:results} shows the mean
squared error between $\log\hat{r}(x|\theta_0, \theta_1)$ and the true
$\log{r}(x|\theta_0, \theta_1)$ (estimated from histograms of $2 \cdot 10^4$
simulations from $\theta_0\!=\! \!-0.8$ and $\theta_1 \!=\! \!-0.6$), summing
over $x \in [5, 15]$, versus the training sample size (which refers to the total
number of $x$ samples, distributed over 10 values of $\theta \in [-1,-0.4]$). We
find that both \scandal and \rascal are dramatically more sample efficient than
pure neural density estimation and the likelihood ratio trick, which do not
leverage the joint score. \rolr improves upon pure neural density estimation and
achieves the same asymptotic error as \scandal, though more slowly.

\paragraph{Lotka-Volterra model.}
As a second example, we consider the Lotka-Volterra
system~\cite{lotka1920analytical, lotka1920undamped}, a common example in the
likelihood-free inference literature. This stochastic Markov jump process models
the dynamics of a species of predators interacting with a species of prey. Four
parameters $\theta$ set the rate of predators and prey being born, predators
dying, and predators eating prey.

We simulate the Lotka-Volterra model with Gillespie's
algorithm~\cite{GILLESPIE1976403}. From the time evolution of the predator and
prey populations we calculate summary statistics $x \in \mathbb{R}^9$. Our model
definitions, summary statistics, and initial conditions exactly follow Appendix
F of Ref.~\cite{NIPS2016_6084}. In addition to the observations, we extract the
joint score as well as the joint likelihood ratio with respect to a reference
hypothesis $\log \theta_1=(-4.61, -0.69, 0.00, -4.61)^T$ from the simulator. On
this augmented data we train different likelihood and likelihood ratio
estimators. As baselines we use \carl~\cite{Cranmer:2015bka, Louppe:2016aov} and
a conditional masked autoregressive flow (MAF)~\cite{2017arXiv170507057P,
repository-maf}. We compare them to the new techniques introduced in
section~\ref{sec:learning_likelihood_ratios}, including a \scandal likelihood
estimator based on a MAF. For MAF and \scandal we stack four masked
autoregressive distribution estimators (MADEs)~\cite{2015arXiv150203509G} on a
mixture of MADEs with 10 components~\cite{2017arXiv170507057P}. For all other
methods, we use three hidden layers. In all cases, the hidden layers have 100
units and $\tanh$ activations. Code for simulation and inference is available at
Ref.~\cite{repository-lotka-volterra}.

For inference on a wide prior in the parameter space, the different probability
densities often do not overlap. As discussed above, the augmented data is then
of limited use. Instead, we focus on the regime where we try to discriminate
between close parameter points with similar predictions for the observables. We
generate training data and evaluate the models in the range $\log (\theta_1)_i -
0.01 \leq \theta_i \leq \log (\theta_1)_i + 0.01$ with a uniform prior in log
space. In the middle panel of Fig.~\ref{fig:results} we evaluate the
different methods by calculating the mean squared error of estimators trained on
small training samples. Since the true likelihood is intractable, we calculate
the error with respect to the median predictions of 10 estimators\footnote{We
pick the algorithms we use for these ``ground truth'' predictions based on the
variance between independent runs and the consistency of improvements with
increasing training sample size. For likelihood estimation, we use the MAF as
baseline, with qualitatively similar results when using \scandal. For likelihood
ratio estimation, we use the \scandal estimator as baseline, and find
qualitatively similar results for \cascal.} trained on the full data set of
200\,000 samples.

Our results indicate a trade-off between the performance in likelihood (density)
estimation and likelihood ratio estimation. For  density estimation, the MAF
performs well. The variance of the score term in the \scandal loss degrades the
performance, especially for larger values of the hyperparameter $\alpha$.
However, for statistical inference the more relevant quantity is the likelihood
ratio. Here the new techniques that use the joint score, in particular \scandal,
are significantly more sample efficient.

\paragraph{Particle physics.}
Finally we consider a real-world problem from particle physics. A simulator
describes the production of a Higgs boson at the Large Hadron Collider
experiments, followed by the decay into four electrons or muons, subsequent
radiation patterns, the interaction with the detector elements, and the
reconstruction procedure. Each recorded collision produces a single
high-dimensional observable $x \in \mathbb{R}^{42}$, and the dataset consists of
multiple iid observations of $x$. The goal is to infer the likelihood of two
parameters $\theta \in [-1, 1]^2$ that characterize the effect of high-energy
physics models on these interactions. We consider a synthetic observed dataset
with 36 iid simulated observations of $x$ drawn from $\theta = (0,0)$.

The new inference techniques can accommodate state-of-the-art simulators, but in
that setting we cannot compare them to the true likelihood function.  We
therefore present a simplified setup and approximate the detector response such
that the true likelihood function is tractable, providing us with a ground truth
to compare the inference techniques to. As simulator we use a combination of
\toolfont{MadGraph~5}~\cite{Alwall:2014hca} and
\toolfont{MadMax}~\cite{Cranmer:2006zs, Plehn:2013paa, Kling:2016lay}. The setup
and the results of this experiment are described at length in
Ref.~\cite{Brehmer:2018eca}, which is attached as supplementary material.

In the right panel of Fig.~\ref{fig:results} we show the expected mean squared
error of the approximate $\log \hat{r}(x | \theta_0, \theta_1)$ for the
different techniques as a function of the training sample size. We take the
expectation over random values of $\theta_0$, drawn from a Gaussian prior with
mean $(0,0)$ and covariance matrix $\diag(0.2^2, 0.2^2)$. We compare the new
techniques to the standard practice in particle physics, in which the likelihood
is estimated through histograms of two hand-picked summary statistics.

All new inference techniques outperform the traditional histogram method,
provided that the training samples are sufficiently large. Using augmented data
substantially decreases the amount of training data required for a good
performance: the \rascal method, which uses both the joint ratio and joint score
information from the simulator, reduces the amount of training data by two
orders of magnitude compared to the \carl technique, which uses only the samples
$x \sim p(x | \theta)$. The particularly simple local techniques \sally and
\sallino need even less data for a good performance. However, their performance
eventually plateaus and does not asymptote to zero error. This is because the
local model approximation breaks down further away from the reference point
$\theta_{\text{ref}} = (0,0)^T$, and the score is no longer the sufficient
statistics. In the supplementary material we show further results and
demonstrate how the improved likelihood ratio estimation leads to better
inference results.

\section{Conclusions}
\label{sec:conclusions}

In this work we have presented a family of new inference techniques for the
setting in which the likelihood is only implicitly defined through a stochastic
generative simulator. The new methods estimate likelihoods or ratios of
likelihoods with data available from the simulator. Most established inference
methods, such as \abc and techniques based on neural density estimation, only
use samples of observables from the simulator. We pointed out that in many cases
the joint likelihood ratio and the joint score\,---\,quantities conditioned on
the latent variables that characterize the trajectory through the data
generation process\,---\,can be extracted from the simulator.

While these additional quantities often require work to be extracted, they also
prove to be very valuable as they can dramatically improve sample efficiency and
quality of inference. Indeed, we have shown that this additional information
lets us define loss functionals that are minimized by the likelihood or
likelihood ratio, which can in turn be used to efficiently guide the training of
neural networks to precisely estimate the likelihood function. A second class of
new techniques is motivated by a local approximation of the likelihood function
around a reference parameter value, where the score vector is the sufficient
statistic. In the case where the simulator provides the joint score, we can use
it to train a precise estimator of the score and use it as locally optimal
summary statistics.

We have demonstrated in three experiments that the new inference techniques let
us precisely estimate likelihood ratios. In turn, this enables parameter
measurements with a higher precision and less training data than with
established methods.

This approach is complementary to many recent advances in likelihood-free
inference: the augmented data can be used to improve training for any neural
density estimator or likelihood ratio estimator, as we have demonstrated for
Masked Autoregressive Flows and \carl. It can also be used to define locally
optimal summary statistics that can be used for instance in \abc techniques.

Finally, these results motivate the development of tools that provide a
nonstandard interpretation of the simulator code and automatically generate the
joint score and joint ratio, building on recent developments in probabilistic
programming and automatic differentiation~\cite{fritz2017, tran2017deep,
siddharth2017learning, le2017inference, gelman2015stan, wood-aistats-2014}. We
have provided a first proof-of-principle implementation of such a tool based on
\toolfont{Pyro}~\cite{repository-automization}.

\subsubsection*{Acknowledgments}

We would like to thank Cyril Becot and Lukas Heinrich, who contributed to this
project at an early stage. We are grateful to Jan-Matthis L\"uckmann for helping
us automate the calculation of the joint likelihood ratio and joint score in
\toolfont{Pyro} and to all participants of the Likelihood-free inference
workshop at the Flatiron Institute for great discussions. We would like to thank
Felix Kling, Tilman Plehn, and Peter Schichtel for providing the
\toolfont{MadMax} code and helping us use it, and to George Papamakarios for
discussing the Masked Autoregressive Flow code with us. KC wants to thank CP3 at
UC Louvain for their hospitality. Finally, we would like to thank At{\i}l{\i}m
G\"{u}ne\c{s} Baydin, Lydia Brenner, Joan Bruna, Kyunghyun Cho, Michael Gill,
Siavash Golkar, Ian Goodfellow, Daniela Huppenkothen, Michael Kagan, Hugo
Larochelle, Yann LeCun, Fabio Maltoni, Jean-Michel Marin, Iain Murray, George
Papamakarios, Duccio Pappadopulo, Dennis Prangle, Rajesh Ranganath, Dustin Tran,
Rost Verkerke, Wouter Verkerke, Max Welling, and Richard Wilkinson for
interesting discussions.

JB, KC, and GL are grateful for the support of the Moore-Sloan data science
environment at NYU. KC and GL were supported through the NSF grants
ACI-1450310 and PHY-1505463. JP was partially supported by the Scientific and
Technological Center of Valpara\'{i}so (CCTVal) under Fondecyt grant BASAL
FB0821. This work was supported in part through the NYU IT High Performance
Computing resources, services, and staff expertise.


\bibliography{references}

\end{document}

%% file: definitions.tex
\newcommand{\toolfont}[1]{\textsc{#1}}

\newcommand{\diff}{\mathrm{d}}
\newcommand{\eg}{{e.\,g.}~}
\newcommand{\ie}{{i.\,e.}~}

\newcommand{\abc}{\textsc{ABC}\xspace}
\newcommand{\nde}{\textsc{NDE}\xspace}
\newcommand{\lrt}{\textsc{LRT}\xspace}
\newcommand{\carl}{\textsc{Carl}\xspace}
\newcommand{\rolr}{\textsc{Rolr}\xspace}
\newcommand{\sally}{\textsc{Sally}\xspace}
\newcommand{\sallino}{\textsc{Sallino}\xspace}
\newcommand{\cascal}{\textsc{Cascal}\xspace}
\newcommand{\rascal}{\textsc{Rascal}\xspace}
\newcommand{\scandal}{\textsc{Scandal}\xspace}

\newcommand{\localmodel}{\ensuremath {{p_{\text{local}}}}}

\newcommand{\intz}{\int \! \diff z\;}

\DeclareMathOperator{\diag}{diag}

\DeclareMathOperator*{\argmin}{arg\,min}

\setlist[itemize]{itemsep=1pt,parsep=1pt, topsep=1pt}